\documentclass[runningheads,a4paper]{llncs}

\usepackage{amssymb}
\usepackage{graphicx}

\usepackage{url}

\begin{document}

\mainmatter  

\title{Architecture of a Web-based Predictive Editor for Controlled Natural Language Processing}

\titlerunning{Architecture of a Web-based Predictive Editor for CNL Processing}

%
%

\author{Stephen Guy \and Rolf Schwitter}

\authorrunning{Stephen Guy \and Rolf Schwitter}

\institute{Department of Computing\\
Macquarie University\\
Sydney, 2109 NSW, Australia \\
\url{{Stephen.Guy | Rolf.Schwitter}@mq.edu.au}}

%
%

\toctitle{Lecture Notes in Computer Science}
\tocauthor{Authors' Instructions}
\maketitle

\begin{abstract}
In this paper, we describe the architecture of a web-based predictive text editor being developed 
for the controlled natural language PENG$^{ASP}$. This controlled language can be used to write 
non-monotonic specifications that have the same expressive power as Answer Set Programs. In 
order to support the writing process of these specifications, the predictive text editor communicates 
asynchronously with the controlled natural language processor that generates lookahead categories
and additional auxiliary information for the author of a specification text. The text editor can display 
multiple sets of lookahead categories simultaneously for different possible sentence completions,
anaphoric expressions, and supports the addition of new content words to the lexicon.

\end{abstract}

\section{Introduction}
Writing a specification in a controlled natural language without any tool support is a difficult task since
the author needs to learn and remember the restrictions of the controlled language. Over the last decade, 
a number of different techniques and tools~\cite{Franconi:11,Fuchs:08,Power:12,Schwitter:03} have been 
proposed and implemented to minimise the learning effort and to support the writing process of controlled 
natural languages. The most promising approach to alleviate these habitability problems is the use of a predictive
text editor~\cite{Schwitter:03,Tennant:83} that constrains what the author can write and provides predictive 
feedback that guides the writing process of the author. In this paper, we present the architecture of a 
web-based predictive text editor being developed for the controlled natural language PENG$^{ASP}$\cite{Schwitter:13}.  
The text editor uses an event-driven Model-View-Controller based architecture to satisfy a number of user
entry and display requirements. These requirements include the display of multiple sets of lookahead
categories for different sentence completions, the deletion of typed words, the addition of new
content words to the lexicon and the handling of anaphoric expressions. Additionally, the text editor displays
a paraphrase for each input sentence and displays the evolving Answer Set Program~\cite{Lifschitz:08}.

\section{Overview of the PENG$^{ASP}$ System}

\subsection{Client-Server Architecture}

The PENG$^{ASP}$  system is based on a client-server architecture where the predictive editor runs 
in a web browser and communicates via an HTTP server with the controlled natural language processor;
the language processor uses in our case an Answer Set Programming (ASP) tool as reasoning service (Fig. 1): 

\vspace{-0.3cm}
\begin{figure}[h!]
\begin{center}
\includegraphics[width=1.0\textwidth]{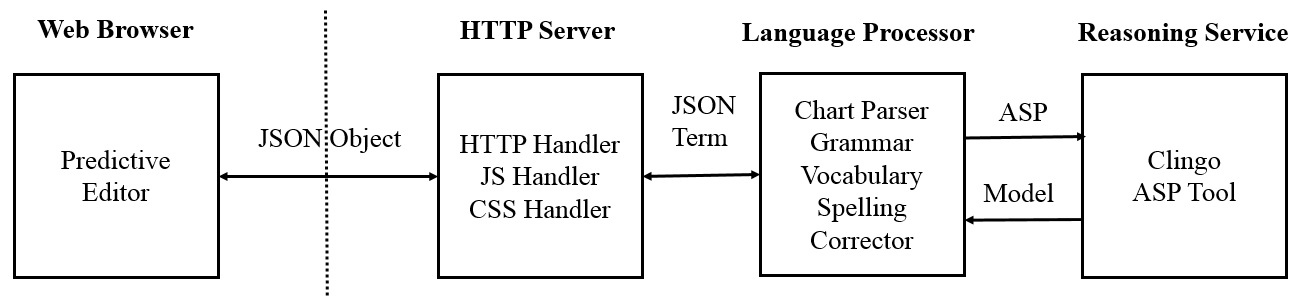}
\caption{Client-Server Architecture of the PENG$^{ASP}$ System}
 \end{center}
\end{figure}
\vspace{-0.5cm}

The communication between the predictive editor and the HTTP server occurs asynchronously with the 
help of AJAX technologies and by means of JSON\footnote{\url{http://json.org/}} objects. The predictive 
editor is implemented in JavaScript\footnote{\url{http://www.ecmascript.org/}} and JQuery\footnote{\url{http://jquery.com/}}.
The HTTP server as well as the controlled natural language processor are implemented in SWI 
Prolog\footnote{\url{http://www.swi-prolog.org/}}. 
The Prolog server translates JSON objects into JSON terms and vice versa so that these terms can be
processed directly by the language processor. The language processor incrementally translates the controlled 
language input via discourse representation structures~\cite{Kamp:93} into an ASP program and sends this ASP program
to the ASP tool {\em clingo}~\cite{Gebser:11,Gebser:12} that tries to generate one or more satisfiable answer sets for the program.

\subsection{HTTP Server}

SWI-Prolog provides a series of libraries for implementing HTTP server capabilities. Our server
is based on this technology and can be operated as a stand-alone server on all platforms that
are supported by SWI-Prolog. The following code fragment illustrates how an HTTP server is
created, a port (\texttt{8085}) specified, and a request (\texttt{Request}) dispatched using a 
handler registration (\texttt{http\_handler/3}):

\begin{verbatim}
   server(Port) :- http_server(http_dispatch, [port(Port)]).
   :- http_handler('/peng/', handle, []).
   handle(Request) :- ...
   :- server(8085).
\end{verbatim}
 
In our case, we can now connect via \texttt{http://localhost:8085/peng/} from the web browser to the
server that uses specific JavaScript and stylesheet handlers to load the predictive editor and to establish
the communication between the editor and the controlled language processor.

\subsection{Predictive Editor}

The predictive editor is implemented in JavaScript and JQuery, with the Superfish\footnote{\url{http://users.tpg.com.au/j_birch/plugins/superfish/}} 
plug-in providing pull-down menu functionality. These technologies allow the 
editor to be run in most browsers, which in conjunction with the capabilities 
of a potentially remote language processor coded in Prolog, provides a highly 
portable system. Data communication with the server provides for both command 
functions, such as file saving and loading, as well as data transfer between the 
language processor and the predictive editor system. The JSON data for parsing sent from the predictive editor to the HTTP server includes the current token 
of a word form, its position in the relevant sentence and relevant sentence number. For 
each word form or completed sentence submitted by the predictive editor, the lookahead 
categories and word forms along with the output of the language processor are returned.

An overview of a typical predictive editor display is presented in Figure 2.
Command function menus are presented at the top, below which is the 
main text input field displaying the current sentence. Lookahead categories for the 
available sentence completion are highlighted using the pull-down menus. Below these lookahead 
categories is a display summarising relevant information in the system, at both the client and server.
First is a summary of previously entered text at the client side.
Second are the generated paraphrases at the server, with any anaphoric references being highlighted
(which may also be accessed from the pull-down menus).
Third is a summary of the current answer set program for the input,
followed by the final section of output from answer set tool {\em clingo}.

\begin{figure}[h!]
\begin{center}
\includegraphics[width=1.0\textwidth]{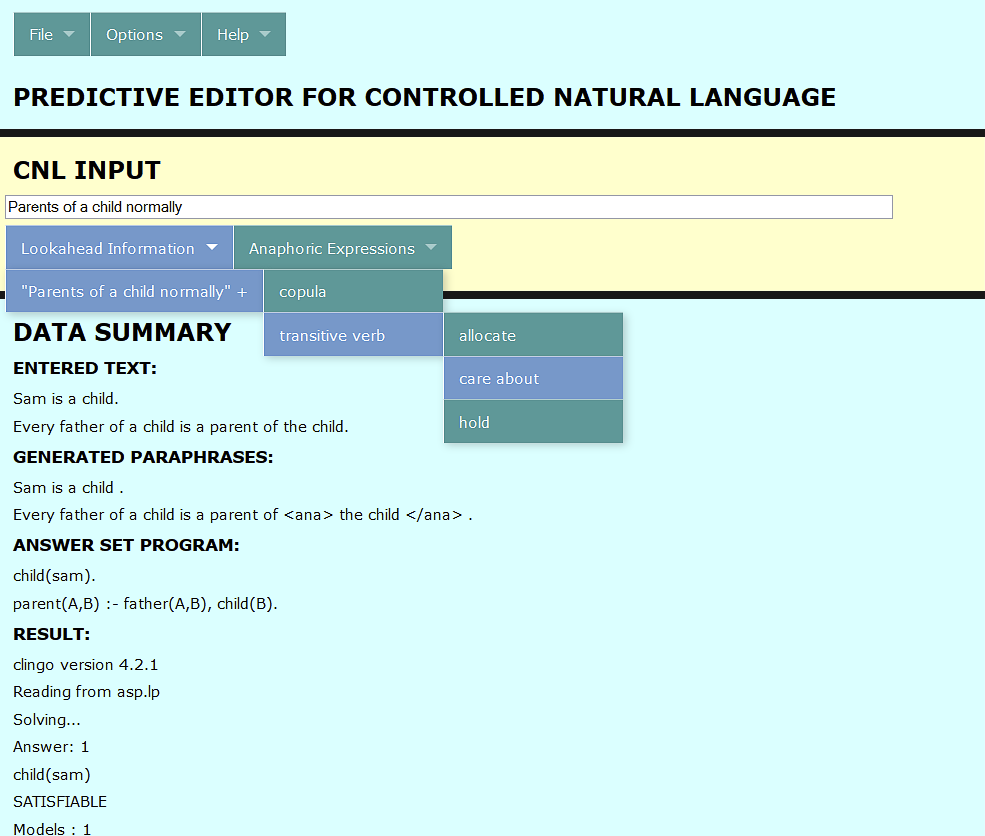}
\caption{Predictive Editor Display}
 \end{center}
\vspace{-0.5cm}
\end{figure}

The editor allows entering text specifications manually by typing in the text entry field, 
plus using pull-down menus of lookahead categories to enter text into the input field. 
The reasons for allowing direct input of text include that some users, especially those 
experienced in the structure of the controlled natural language, can type faster than 
they can enter via menus, even with some level of auto-completion. Additionally, the system allows entering new content words into 
the lexicon, via the text field, that do not appear in the displayed lookahead categories.

\section{Processing and Reasoning in the PENG$^{ASP}$ System}

\subsection{Controlled Natural Language Processor}

The controlled natural language processor of the PENG$^{ASP}$ system consists 
of a chart parser, a unification-based grammar, a lexicon and a spelling corrector. The chart 
parser is initialised for the first time when the author moves the cursor into the textfield of the 
predictive editor and reset at the beginning of each new sentence and generates lookahead 
categories using the grammar and the lexicon of the controlled language processor. These 
lookahead categories inform the author of a specification how to start a sentence and are generated 
dynamically for each word form that the author enters into the textfield of the editor. This mechanism 
guarantees that the author can only input word forms and construct sentences that follow the rules of 
the controlled language. If a word is misspelled, then the spelling corrector is used to generate a list 
of candidates that occur in the lexicon. If a content word is not in the lexicon, then the author can
add this word to the lexcion during the specification process.

The controlled natural language PENG$^{ASP}$~\cite{Schwitter:13} that the author uses as input language
has been designed as a high-level interface language to ASP programs. In certain aspects the language PENG$^{ASP}$ is
similar to PENG Light~\cite{White:09} and Attempto Controlled English~\cite{Fuchs:08}, since it
uses a version of discourse representation theory (DRT), in the spirit of~\cite{Eijck:11,Kamp:93},
as intermediate representation language. However, PENG$^{ASP}$ does not rely on full 
first-order logic (FOL) as target language as the use of DRT would suggest but on the language 
for ASP programs. The language of FOL is in some respects more expressive than the language of ASP but
unfortunately FOL is not adequate for representing commonsense knowledge, because FOL cannot
deal with non-monotonic reasoning. ASP, on the other hand, allows us to represent and process commonsense
knowledge because of its unique connectives and non-monotonic entailment relation. Beyond that, 
ASP is still expressive enough to represent function-free FOL formulas of the
 \small\texttt{$\exists^{*}\forall^{*}$} \normalsize prefix class in form of a logic 
program~\cite{Lierler:13}.  Below is an example specification in PENG$^{ASP}$ that uses a
default rule in (5), a cancellation axiom in (6), and sentence with strong negation in (7):

\begin{itemize}

\item[1.] Sam is a child.

\item[2.] John is the father of Sam and Alice is the mother of Sam.

\item[3.]  Every father of a child is a parent of the child.

\item[4.]  Every mother of a child is a parent of the child.

\item[5.]  Parents of a child normally care about the child.

\item[6.]  If a parent of a child is provably absent then the parent abnormally cares about the child.

\item[7.]  John does not care about Sam.

\item[8.]  Alice is absent.

\end{itemize}

Of course, the specific features of the ASP language have an impact 
on what we can express on the level of the controlled natural language and therefore rely on the support
of the predictive editor.

\subsection{Reasoning Service}

Since we are interested in specifying commonsense theories in PENG$^{ASP}$, we need a non-monotonic reasoning service.
ASP is a relatively novel logic-based knowledge representation formalism that has its roots in logic programming 
with negation, deductive databases, non-monotonic reasoning and constraint solving~\cite{Brewka:11,Gebser:12}. An ASP 
program consists of a set of rules of the following form:


\begin{itemize}
\item[]  \texttt{ L$_{0}$ ; ... ;  L$_{k}$  :-  L$_{k+1}$,  ...,  L$_{m}$,  not  L$_{m+1}$,  ..., not  L$_{n}$}.
\end{itemize}

\noindent where all \texttt{L$_{i}$}'s are literals. A literal is an atom or its negation. A positive atom has the form \texttt{p(t$_{1}$, ..., t$_{n}$)} 
where \texttt{p} is a predicate symbol of arity \texttt{n} and \texttt{t$_{1}$, ..., t$_{n}$} are object constants or variables. A negative atom
has the form \texttt{-p(t$_{1}$, ..., t$_{n}$)} where the symbol \texttt{-} denotes strong negation.
 The symbol \texttt{:-} stands for
an implication. The expression on the left-hand side of the implication is called the {\em head} of the rule and the expression on the 
right-hand side is called the {\em body} of the rule. The head may consist of an epistemic disjunction of literals denoted by the symbol
\texttt{;}. Literals in the body may be preceded by negation as failure denoted by the symbol \texttt{not}.  The head or the body of a rule 
can be empty. A rule with an empty head is called an {\em integrity constraint} and a rule with an empty body is called a {\em fact}. 
For instance, the example specification in Section 3.1 is translated automatically via discourse representation structures in the
subsequent ASP program: 

\begin{verbatim}
   child(sam).
   father(john,sam).
   mother(alice,sam).
   parent(A,B) :- father(A,B), child(B).
   parent(C,D) :- mother(C,D), child(D).
   care(E,F) :- parent(E,F), child(F), not ab(d_care(E,F)), 
                not -care(E,F).
   ab(d_care(G,H)) :- parent(G,H), child(H), not -absent(G).
   -care(john,sam).
   absent(alice).
\end{verbatim}

\section{Predictive Editor Requirements}

In addition to the generic requirements outlined in Section 2.3, a number of detailed user input and system display requirements for the lookahead categories are determined to aid in the design of the predictive editor architecture. The main requirements are that the system should allow appropriate editing of information already entered, that the lookahead categories for a particular sentence position are displayed until all possibilities are no longer possible and that the lookahead categories for the next sentence position are displayed as soon as the relevant options are possible. These requirements are presented in detail in the following
sections.

\subsection{User and System Requirements} 

\subsubsection {User Entry Requirements}

\begin{itemize}
\item[] {\bf Requirement E.1.1:} The system will allow deletion of characters or words already typed, or all or part of a sentence not yet {\em submitted}. (This deletion will be referred to as {\em backward editing}).
\end{itemize}

\begin{itemize}
\item[] {\bf Requirement E.2.1:} A new sentence is not commenced (via the chart parser being reset) until a {\em submit} or an {\em enter} event or a beginning of sentence character/word occurs after an end-of-sentence marker (full stop or question mark). A new sentence being commenced means that the previous sentence has been {\em submitted}.
\end{itemize}

\begin{itemize}
\item[] {\bf Requirement E.3.1:} A user is allowed to enter a content word not in the lexicon and force its submission to the language processor as the next content word.
\end{itemize}

\begin{itemize}
\item[] {\bf Requirement E.3.2:} A user may enter a misspelt word that is yet to be {\em completed} with the word still subject to {\em backward editing}.
\end{itemize}

\begin{itemize}
\item[] {\bf Requirement E.4.1:} A word is {\em completed} if it followed by a space or directly by a valid punctuation character which in turn is followed by a space or sentence {\em submission}. This latter requirement of a space after the punctuation allows the system to distinguish the state from the case of an {\em incomplete} misspelt word with an erroneous punctuation character at the end.
\end{itemize}

\subsubsection {System Display Requirements}

\begin{itemize}
\item[] {\bf Requirement D.1.1:} Before and whilst a word is being entered at position A (or for a new sentence commencing at position A), the system should display all the lookahead categories for position A until all of those categories are no longer possible.
\end{itemize}

\begin{itemize}
\item[] {\bf Assertion D.1.1:} All lookahead categories for position A are no longer possible if the next non-punctuation word at position A+1 has commenced, or a word is {\em completed} according to Requirement E.4.1.
\end{itemize}

\begin{itemize}
\item[] {\bf Requirement D.2.1:} The system should display the lookahead categories for position A+1 when a word entered at position A matches the lookahead categories for position A.
\end{itemize}

Note that in terms of displaying one set of lookahead categories for a particular word, requirements D.1.1 and D.2.1 are not mutually exclusive, that is there occur system states where the lookahead categories at position A and position A+1 need to be displayed concurrently.

\begin{itemize}
\item[]  {\bf Assertion D.2.1:} If a word at position A matches the lookahead categories for position A, then other lookahead categories for position A may still be possible.
\end{itemize}

\subsection{Display of Multiple Sentence Completions}

Some examples are presented to help clarify the requirements detailed above. The two main cases which are catered for are the existence of subsets within the lookahead categories for one sentence position and the allowed juxtaposition of punctuation directly after a word without an intervening space.

For the case of subsets in lookahead categories, consider the commencement of a sentence and the above two display requirements D.1.1 and D.2.1. Initial lookahead categories may include ``The", ``There is", ``A", ``Thelma", ``John" and ``Johnathan" for example, which according to D.1.1 should all be displayed by the system. A user entering the characters ``The" would then satisfy requirement D.2.1, whereby the lookahead categories for the next position would be displayed. If these categories included the word ``child", the user could enter this word and the entered text would be ``The child", illustrating that a display of this sentence completion option was necessary. However, the original situation of the user entering the characters ``The" may have been the precursor to the entry of the words ``There is" or even ``Thelma". Thus even though requirement D.2.1 is satisfied after the entry of ``The", requirement D.1.1 still holds for the presentation of the original lookahead categories whilst the user completes this entry, thus illustrating assertion D.2.1. Whether the user has entered ``Thelma" or ``The" without a subsequent character, requirement E.4.1 has not been satisfied, so a user may {\em backward edit} from the word ``Thelma" back to ``The" or ``Thelma"/``The" back to ``A".

For the case of juxtaposition of word forms with punctuation and requirements E.4.1 and D.2.1, the lexicon and grammar allows phrases such as ``John, Thelma and Pete are parents.". Here, a word is followed directly by punctuation, so that once the characters ``John" are entered, according to requirement D.2.1, the system must display the options for the next lookahead categories which include the comma which could be clicked or typed directly. Alternatively, a user may have been intending to type ``Johnathan", so as for the case of subsets must see the original set of lookahead categories. If a user accidentally hit the comma on the fifth character, leaving ``John," (John comma), as the current word, the system should still display the original lookahead categories, including ``Johnathan", as the word has not been {\em completed} according to requirement E.4.1.

\section{Architecture of the Predictive Editor}

The predictive editor is designed to meet the requirements of the PENG$^{ASP}$ system, the 
asynchronous client-server communications, the different modes of the editor input as well as user entry and system display requirements.

\subsection{Model-View-Controller Architecture}

The architecture of the predictive editor is based approximately on that of a
 {\em Model-View-Controller} (MVC) system~\cite{Freeman:04,Sommerville:11} in terms of separation and independence.

The {\em Model} includes the currently active sentence, including that entered by the user and that submitted 
to the HTTP server, all previously entered sentences and all data (including lookahead categories) received from 
the language processor via the HTTP server. The model also stores all variables relevant to determining the state 
of the system.

 The {\em View} includes the events-triggered input text field, the pull-down menu display of 
lookahead categories and the input of word forms via mouseover selection. It also displays the overall model 
of entered sentences and the ASP model generated by the language processor. 

The {\em Controller} synchronises 
all functions, and importantly monitors for the need of a state change in the {\em Model}, such as when the user has input data that is different from the currently active sentence and if so, whether to submit new data to the server or not. Additionally, the {\em Controller} 
co-ordinates loading of all the returned lookahead categories into data structures and determines which of these 
lookahead categories are displayed to the user as dependent on the current state of the system.

\subsection{Event-Triggered Implementation}

A key issue with the implementation of the MVC architecture is the requirement to have event-driven data processing 
and control to be compatible with the asynchronous AJAX communication between the predictive editor and the 
HTTP server and events-triggered predictive editor input. When content words are submitted to the HTTP server 
via JSON data, the predictive editor system must wait until corresponding lookahead data is returned by the server. 

Once this information is received, it may then be stored in the model and only then can the {\em Controller} process 
this model data to determine if the model state variables should be changed and update the display if necessary. To 
implement this, the {\em Controller} organises run-time execution of events in a pipe and filter architecture, where 
each element of the pipe is a data structure containing the relevant primary data for that event, the relevant processing 
function and an optional link to the next data structure in the pipe.  

Whilst this may not be a classical MVC implementation, it provides a robust method of ensuring model data is in a consistent
 state for process control. Thus for the above example of sending a new content word to the server, the AJAX send/receive routine 
will trigger the return data storage event, which when complete will trigger the model state change assessment functionality, 
which when complete may cause a trigger of the display of the next lookahead categories to the display. 

Any multi-stage data processing may also be organised as a pipe and filter structure using the above data structures, with the 
next stage of the processing function only allowed once the model data from the previous processing function becomes stable.

\subsection{Data Structures}

As with many client-server systems, some model data is stored and processed at the predictive editor client side to 
allow for optimal processing and control. The model data is stored in objects defined by JavaScript functions, with 
appropriate object methods declared to allow for this data to be processed conveniently and allowing functionality 
beyond the capabilities of using raw JSON objects for storage. For example, the model data includes stack objects 
(containing stacks of anything from word forms to whole sentences), individual send and received objects plus a 
single object of correlated send and receive data. Methods can detect if a beginning or end of sentence token is 
present, or whether a word form matches a lookahead category and whether it is also a subset of another lookahead 
category (such as ``The" being a subset of ``There"). Display objects allow storage of different sets of lookahead 
categories and the ability to switch the display from `displayed' to `hidden' and vice versa.

\subsection{Predictive Editor Controller}

Given the user entry and system display requirements discussed Section 4.1 and generic requirements presented in Section 2.3, the control system for the predictive editor has been designed to allow displaying of multiple lookahead categories for different sentence completions and strict control over when data entered by a user is ultimately committed to the server. The currently active sentence is stored in two forms, namely from a tokenisation of the user input and from a summary of the data submitted to the server. By comparing a stack of the set of tokens in each sentence, a difference stack is generated to aid the controller in determining a change in the model state. Any newly entered valid words, or changes in the current word are assessed for submission, or alternately earlier submitted tokens/words may be removed and new tokens sent in their place (such as in the case of {\em backward editing}). 

As discussed regarding requirement D.2.1 in Section 4.1, if an entered word matches a lookahead category for that position, the controller automatically submits this word to the server and retrieves the next set of lookahead categories for this new token. However, this data transfer is just the predictive editor gathering information and doesn't directly synchronise with the totality of the display to the user. If the controller doesn't detect a word completion, or finds that at least one lookahead category from the previous word is still possible, the previous lookahead categories are not cleared as per assertion D.2.1. 

As described in Section 5.3, display data structures allow easy addition and display of data and hiding of data as necessary. As well as automatically submitting a word matching the current lookahead categories, a word matching the previous set of lookahead categories where the previous word is a subset of the new word will also trigger an automatic submission of the token to the HTTP server. This would be the case for ``Thelma" being typed after ``The" has been submitted to the server and lookahead categories already returned for the next sentence position.

\subsection{Adding Content Words to the Lexicon}

Recall from requirement E.3.1 that a user may forcibly submit a word form to the language processor that does not correspond to the lexicon. When this occurs, the language processor may offer a set of spelling suggestions (assuming that an incorrect word has been submitted by mistake) or the predictive editor will offer an option to add this new word to the lexicon in this current context. If the user selects to add a word, then the position in the sentence, the lexical category and the new word form are collected and sent to the server where the new word is added to the lexicon. The new word is then parsed again by the language processor and a new set of lookahead categories is generated and sent to the predictive editor. 

\section{Future Research}

The current predictive editor may be extended for multiple users in line with the web-based portability of the system. A user login would allow for a number of features, such as a user-group based lexicon depending on the nature of the specification system for that group (e.g. medical, engineering, automotive, etc.). Additionally, an individual could have their own extended lexicon for any content words added to the lexicon. A user could set a level of knowledge for their grammar, which would aid in controlling the complexity of the pull-down menus, in that instead of displaying all possible lexical categories, a user with limited knowledge could display a smaller number of less-technial word categories, such as ``function words" instead of individual groups such as ``adjective", ``adverb", ``noun", etc. The user login could be used to set preferences for any further adjustable enhancements.

\section{Conclusion}

In this paper, we introduced the architecture of a web-based predictive text editor developed for the PENG$^{ASP}$ system. 
This system is suitable for writing non-monotonic specifications that have the expressive power of Answer Set
Programs. The web-based predictive editor supports the writing process of these specifications and is based on a portable
client-server architecture and is predominantly implemented in JavaScript. An event-driven Model-View-Controller based 
architecture was used for the editor, allowing strict control of system functionality to satisfy a set of user entry and display 
requirements that included the display of multiple sets of lookahead categories for different sentence completions.
The predictive editor allows for new content words to be added to the lexicon and supports the selection of anaphoric expressions
An extension of a user login would allow tailoring of preferences and a user-based lexicon.

\end{document}